\documentclass[runningheads,a4paper]{llncs}

\usepackage{amssymb}
\usepackage{graphicx}
\usepackage{caption}
\usepackage{amsmath, mathpazo}
\usepackage{amssymb}
\usepackage{float}
\usepackage{stackengine}
 
\usepackage[style=numeric,backend=bibtex]{biblatex} 
\bibliography{bookchapter}

\usepackage{url}
\urldef{\mailsa}\path|{ali-boyali, smita}@toyota-ti.ac.jp|    
\newcommand{\keywords}[1]{\par\addvspace\baselineskip
\noindent\keywordname\enspace\ignorespaces#1}

\begin{document}

\mainmatter  

\title{A Tutorial On Autonomous Vehicle Steering Controller Design, Simulation and Implementation}

\titlerunning{A Tutorial in Motion Controllers}

%
%
\author{Ali Boyali, Seichi Mita 
\and Vijay John}
%

\institute{Toyota Technological Institute,\\
468-8511 Aichi Prefecture, Nagoya, Tenpaku Ward, Hisakata, 2 Chome 1-2-1 \\
\mailsa\\
\url{http://www.toyota-ti.ac.jp/english/}}

%
%

\tocauthor{Authors' Instructions}
\maketitle

\begin{abstract}
This tutorial details simple low-level lateral motion controllers for self-driving car path following and provides practical methods for curvature computation given the recorded vehicle path data.  

\keywords{Self-driving car steering controllers, autonomous parking, curvature computation methods}
\end{abstract}

\section{Introduction}

The global players in the automotive industry have been trying to launch fully autonomous vehicles into the mass market shortly. This tutorial presents the basic information required to design motion controllers for autonomous vehicles with simplified details.  

\section{Vehicle Models}
\begin{figure}[H]
  \centering	
  \includegraphics[scale=0.5]{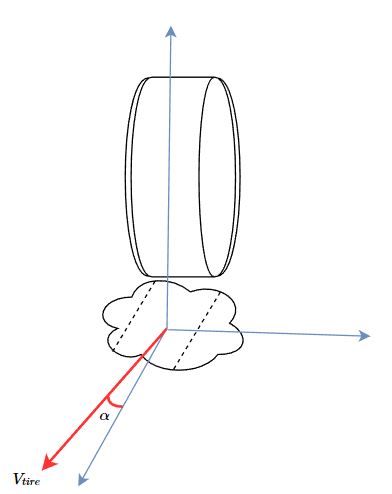}
  \captionsetup{justification=centering}
  \caption{Kinematic Vehicle Model}
  \label{fig:tire}
\end{figure}

Tires are the main components of road vehicles that provide guidance and force generation on roads. As such, they define motion characteristics, dynamics and kinematics of road vehicles. Due to its elastic structure, tire motion characteristics are nonlinear and complex. The velocity vector of tires at the tire-road contact patch centre points a different direction than the tire heading direction at high speeds. The deviation of the velocity vector direction from the tire vertical plane is called tire sideslip angle (Figure \ref{fig:tire}). 
 
The tire sideslip angle is negligible at low speeds, and kinematic equations can approximate the vehicle's motion. The kinematic models are used to design controllers for low-speed maneuvers such as tight parking and, in some cases, for motion planning. Dynamical vehicle models are necessary for high-speed vehicle motion as required in urban and highway driving. In this tutorial, we derive vehicle lateral motion controllers using dynamic and kinematic modelling approaches and present the corresponding low-level lateral controllers from a practical implementation perspective.
  
\subsection{Kinematic Models}

The systems with less actuation direction than the available motion coordinates in the configuration space are called non-holonomic (non-integrable) systems. On the vehicle motion plane in the global coordinate system, the velocities of the front and rear axle centres are constrained, and they are not independent of each other. 

\begin{figure}[H]
  \centering	
  \includegraphics[scale=0.5]{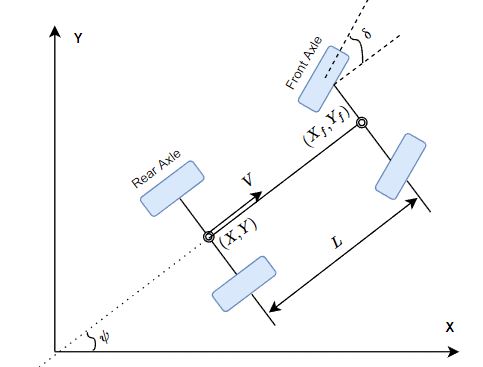}
  \captionsetup{justification=centering}
  \caption{Kinematic Vehicle Model}
  \label{fig:kinmodel}
\end{figure}

Two points on the rear and front axles of the vehicle are shown in (Figure \ref{fig:kinmodel}). The vehicle body frame rigidly connects these points. The kinematic models are derived for the chosen point and may show variations in the equations. These points are chosen to some reference tracking requirements. For example, in parking maneuvers, the rear axle centre is required to track the reference path, especially in the reverse motion.  

The velocities of the rear and front axle centres are constrained by the following forms \cite{latombe2012robot};
 
\begin{align}
\begin{split}\label{eq:1}
    tan(\psi) = \frac{\dot{Y}}{\dot{X}} \implies & \dot{X}sin(\psi)-\dot{Y}cos(\psi) = 0    
\end{split}\\
\begin{split}\label{eq:2}
  tan(\psi+\delta) = \frac{\dot{Y}}{\dot{X}} \implies & \dot{X}sin(\psi+\delta)-\dot{Y}cos(\psi+\delta) = 0
\end{split}  
\end{align}

In this work, we take the rear axle centre as the reference tracking point. The kinematic model is constructed with respect to this reference point. In this case, the front axle centre velocity can be described by the rear axle centre velocity, yielding the following constraint relationship. These velocity constraint equations are commonly called the Pfaffian constraint equations \cite{latombe2012robot, murray1994mathematical}. 

\begin{align}
  \begin{split}\label{eq:3}
      X_f = X + Lcos(\psi)  
  \end{split}\\  
  \begin{split}\label{eq:4}
      Y_f = Y + Lsin(\psi)  
  \end{split}
\end{align}

Using the angle-sum identity in trigonometry in Equation \ref{eq:2} and substituting the Equations (\ref{eq:3}) and (\ref{eq:4}), we arrive at the Pfaffian constraint matrix and the kinematic equations of motion \cite{latombe2012robot}.

Substituting $X_f= X + Lcos(\psi) $ and $Y_f = Y + Lsin(\psi)$ in Equation (\ref{eq:2}) results in;

\begin{equation}
	\label{eq:5}
	\dot{X}sin(\psi+\delta)-\dot{Y}cos(\psi+\delta)-\dot{\psi}Lcos(\delta) = 0    
\end{equation}

\begin{equation}
\label{eq:6}
C(q)= 
	\begin{bmatrix}
    	sin(\psi+\delta) & -cos(\psi+\delta) & -Lcos(\delta)\\
        sin(\psi) &  -cos(\psi) & 0
	\end{bmatrix}
\end{equation}

The rate of heading angle can be obtained from Equation (\ref{eq:5}). Therefore, the kinematic differential equations for the configuration space $\{X,{\:} Y,{\:} \psi \} \in S$ that describe the motion in the global coordinate system become;

\begin{equation}
\label{eq:7}
\begin{bmatrix}
	\dot{X}\\ \dot{Y} \\ \dot{\psi}
\end{bmatrix}
= 
	\begin{bmatrix}
    	Vcos(\psi)\\
        Vsin(\psi) \\ 
        \frac{V}{L}tan(\delta)
	\end{bmatrix}
\end{equation}

The control variables determine the longitudinal and lateral motions of the vehicle, vehicle velocity $V$ and the steering angle $\delta$ of the tires. Given the control inputs $(V, \: \delta)$, the motion trajectories of the vehicle can be simulated.

In the controller design, a feedback control law for lateral and longitudinal motions can be developed separately, and however since the kinematic equations are non-holonomic, smooth feedback cannot be determined for the point-to-point control applications in which two-point boundary conditions are to be satisfied for all coordinates in the configuration space. We detail some practical controller design methods in the proceeding sections.

\subsection{Dynamic Models} 

\begin{figure}[H]
  \centering	
  \includegraphics[width=1.0\textwidth]{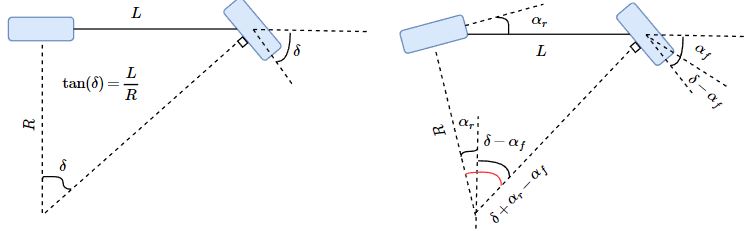}
  \captionsetup{justification=centering}
  \caption{Ackerman Steering Geometry (left) and Side-slip Angles (left)}
  \label{fig:ackermann}
\end{figure}

We briefly introduced the tire sideslip angle that generates lateral and longitudinal tire forces in the previous section. Tire slip angle cannot be neglected after after some velocity values.  These forces are expressed as a function of the tire deflection. At these conditions, the kinematic relations are no longer valid for Ackermann steering geometry (left Figure in \ref{fig:ackermann}).  

In Fig. (\ref{fig:ackermann}), the Ackerman steering geometry is depicted on the right. The steering angle agrees with the curvature, which can be formulated using a trigonometric relationship. The tangent of the steering angle is equal to the ratio of the length of the vehicle and the radius of curvature.  

\begin{figure}[H]
  \centering	
  \includegraphics[width=0.7\textwidth]{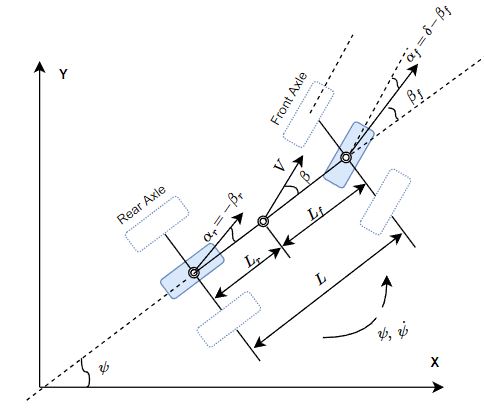}
  \captionsetup{justification=centering}
  \caption{Dynamic Vehicle Model reduced to Single Track}
  \label{fig:singletrack}
\end{figure}

In the derivation of the lateral vehicle dynamics equations, the common approach uses a single-track model (Figure \ref{fig:singletrack}). In this approach, the four-wheel model is lumped into the two-wheel single track structure \cite{rajamani2011vehicle}.

We made the following assumptions in derivation of the single-track model:
\begin{itemize}
\item Side slip and steering angles are small for linearization,
\item tires operate at the linear region in which the slope of tire slip angle and lateral force curve is constant,
\item the road surface and tire friction coefficient $\mu$ is constant (we omit friction coefficient by taking $\mu=1$),  
\item the vehicle travels at a constant longitudinal speed.
\end{itemize}

The net lateral force acting at the centre of gravity of the vehicle in the body coordinate system is written using Newton's second law of motion ;

\begin{align}
  \begin{split}\label{eq:8}
     \sum{F_y} =ma_y = F_{yf} + F_{yr}  
  \end{split}\\  
  \begin{split}\label{eq:9}
      F_{yf} = 2C_{{\alpha}f}\alpha_{f}
  \end{split} \\
  \begin{split}\label{eq:10}
      F_{yr} = 2C_{{\alpha}r}\alpha_{r} 
  \end{split}
\end{align}

where $C_{\alpha_f}, \: C_{\alpha_f}$ are the cornering stiffness of the front and rear tires for per angle deviation in radian and $\beta$ is the sideslip angle of the vehicle at the Center of Gravity (CoG). The sideslip angles of the vehicle at the front and rear axle centres $(\beta_f, \: \beta_r)$ are computed from the kinematic velocity relationships and given as;

\begin{align}
  \begin{split}\label{eq:11}
     \beta =atan(\frac{\dot{y}}{V_x}){\approx} \frac{\dot{y}}{V_x}
  \end{split}\\  
  \begin{split}\label{eq:12}
      \beta_{f} \approx\frac{\dot{y}+L_f\dot{\psi}}{V_x}
  \end{split} \\
  \begin{split}\label{eq:13}
       \beta_{f} \approx\frac{\dot{y}-L_r\dot{\psi}}{V_x} 
  \end{split}
\end{align}

Substituting these identities in the lateral force balance equation (\ref{eq:8}) yields the lateral motion differential equation;

\begin{equation}
\label{eq:14}
	\frac{d}{dt}\dot{y} = -\frac{2(C_{{\alpha}f}+C_{{\alpha}r})}{mV_x}\dot{y}-V_x\dot{\psi}-\frac{2(C_{{\alpha}f}L_f-C_{{\alpha}r}L_r)}{mV_x}\dot{\psi}+\frac{2C_{{\alpha}f}}{m}\delta
\end{equation}

Similarly, we write the net moment balance equation (\ref{eq:15}) and derive the yaw rate update equation (\ref{eq:16}). These differential equations are used to obtain the update equations for lateral displacement and the heading angle by integrating the rates and can be put in matrix form for state-space representation. 

\begin{equation}
    \label{eq:15}
	I_z\ddot{\psi} = F_{yf}L_f - F_{yr}L_r 
\end{equation}

\begin{equation}
	\label{eq:16}
	\frac{d}{dt}\dot{\psi} = -\frac{2({L_f}C_{{\alpha}f}-{L_r}C_{{\alpha}r})}{{I_z}V_x}\dot{y}-\frac{2(L_f^2C_{{\alpha}f}+L_r^2C_{{\alpha}r})}{{I_z}V_x}\dot{\psi}+\frac{2{L_f}C_{{\alpha}f}}{I_z}\delta
\end{equation}

The state space representation of the lateral dynamics can be written for the vehicle state $\{y, \:\dot{y}, \:\psi, \:\dot{\psi}\} $;

\begin{equation}
\label{eq:17}
\frac{d}{dt}
	\begin{bmatrix}
	y \\ \dot{y} \\ \psi \\ \dot{\psi}
	\end{bmatrix} = 
    \begin{bmatrix}
    0 & 1 & 0 & 0 \\
    0 &-\frac{2(C_{{\alpha}f}+C_{{\alpha}r})}{mV_x} &0 &-V_x-\frac{2(C_{{\alpha}f}L_f-C_{{\alpha}r}L_r)}{mV_x}\\
    0 & 0 & 1 & 0 \\
    0 & -\frac{2({L_f}C_{{\alpha}f}-{L_r}C_{{\alpha}r})}{{I_z}V_x}& 0 &-\frac{2(L_f^2C_{{\alpha}f}+L_r^2C_{{\alpha}r})}{{I_z}V_x}
    \end{bmatrix}
    \begin{bmatrix}
	y \\ \dot{y} \\ \psi \\ \dot{\psi}
	\end{bmatrix}+
    \begin{bmatrix}
	0 \\ \frac{2C_{{\alpha}f}}{m}\\ 0 \\ \frac{2{L_f}C_{{\alpha}f}}{I_z}
	\end{bmatrix} \delta
\end{equation}

The appropriate units are accordingly in the equations. 

\section{Controller Design} 

In a control application, the main objective is to find a control input sequence that brings the system output closer to predefined reference trajectories. If the system's output is required to be zero all the time, the problem is called a regulation problem where the reference is zero. We presented the kinematic and dynamic motion equations in the previous sections for the specific configuration spaces. The motion equations can be redefined in the error space to transform the control design into the regulation problem to bring the non-zero system outputs and their derivatives to zero. Once the error equations are derived, we can design feedback controllers that take the error states as input and give the required amount of the control input magnitude. It is paramount to note a distinction in the control objective here. The control objectives differ in the number of regulated output vs the number of controlled input variables that might incur difficulties in finding a smooth feedback law. Three types of control objectives are defined in \cite{latombe2012robot}. These are;

\begin{itemize}
  \item \textbf{Point-to-Point control}; Given an initial state configuration, we desire the vehicle to reach a goal configuration in which all the position requirements are satisfied. As an example, parking control can be classified under this control task. The objective is to find controller inputs (two inputs for a vehicle; speed and steering) to satisfy the end-point position described by $\{x,  \:y, \:\psi\}$. If the vehicle speed is assumed to be constant, the parking control becomes a one-input three-output control problem for which there is no smooth feedback solution.
  
  \item \textbf{Path following}; the vehicle is desired to follow a geometric path from a given initial position. If the vehicle speed is taken as constant, the path following becomes one input, one output control problem for which various controllers can be developed for smooth feedback. In this case, the deviation of the vehicle position from the defined path is regulated.
  
  \item \textbf{Trajectory following}; the vehicle is desired to follow a trajectory which is a function of time. The optimal trajectory is generated by the open-loop optimization method, and feedback laws are designed to keep the vehicle on the trajectory. 
\end{itemize}
 
\subsection{Optimization-Based Kinematic Control}

We derive a steering control law for the kinematic model, assuming that the vehicle speed is not a control variable. The only free variable for control is the steering input to the system. Recall that the kinematic equations are given for the two-axle vehicle system are given as; 

\begin{equation}
\label{eq:18}
\begin{bmatrix}
	\dot{X}\\ \dot{Y} \\ \dot{\psi}
\end{bmatrix}
= 
	\begin{bmatrix}
    	Vcos(\psi)\\
        Vsin(\psi) \\ 
        \frac{V}{L}tan(\delta)
	\end{bmatrix}
\end{equation}

If we were to design a feedback controller, the control design task would fall under the trajectory following problem. However, in this paper, we will use the $x(t)$ position as the index value that gives us the other configuration space variables $y(t) \: and \:\psi(t)$ at any time. Other assumptions are that the trajectory is generated by a trajectory generator or a feasible trajectory is available and the vehicle travels at a constant speed. With these assumptions, we can reduce the trajectory tracking problem into the path following regulation for the lateral distance and heading deviations. We start with linearization of the kinematic equations around a fixed point $(s_*, u_*)=\{x_*, \:y_*,\:\psi_*, \:\delta_*\}$ to obtain the linearized model by the first-order Taylor expansion \cite{murray2009optimization}. 

\begin{align}
  \begin{split}
      \dot{s}=f(s, u)
  \end{split}\\
  \begin{split}
  \label{eq:21}
      \dot{s}\:{\approx}\:f(s_*, u_*) + \frac{\partial f(s, u)}{\partial s}\Bigr|_{\substack{s=s_*\\u=u_*}}(s-s_*)+\frac{\partial f(s, u)}{\partial u}\Bigr|_{\substack{s=s_*\\u=u_*}}(u-u_*)
  \end{split}	
\end{align}

By subtracting $\dot{s_*}=f(s_*, u_*)$ from Equation (\ref{eq:21}), we arrive the linearized equations in relative coordinates;

\begin{equation}
	\dot{s_e}=\dot{s-s_*}\:{\approx}\: \frac{\partial f(s, u)}{\partial s}\Bigr|_{\substack{s=s_*\\u=u_*}}(s-s_*)+\frac{\partial f(s, u)}{\partial u}\Bigr|_{\substack{s=s_*\\u=u_*}}(u-u_*)
\end{equation}

where $s_e$ is the states in a relative coordinate system and $u$ is the control variable. Before deriving the Jacobian matrices, we rewrite the kinematic model equations for the desired trajectory in the vehicle body coordinate system. The kinematic model for the desired trajectory in the global coordinate system is;

\begin{equation}
\begin{bmatrix}
	\dot{X}\\ \dot{Y}
\end{bmatrix}
= 
	\begin{bmatrix}
    	Vcos(\psi_d)\\
        Vsin(\psi_d) 
	\end{bmatrix}
\end{equation}

The exact identities in the local coordinate system can be written by incorporating the rotation matrix into the equation and re-arranging the trigonometric identities. Therefore, desired path equations in the local coordinate system become;

\begin{equation}
\label{eq:23}
\begin{bmatrix}
	\dot{x}\\ \dot{y} 
\end{bmatrix}
= \begin{bmatrix}
	cos(\psi) & sin(\psi) \\-sin(\psi) & cos(\psi)
  \end{bmatrix}
	\begin{bmatrix}
    	Vcos(\psi_d)\\
        Vsin(\psi_d)
	\end{bmatrix}=	\begin{bmatrix}
                        Vcos(\psi-\psi_d)\\
                        Vsin(\psi-\psi_d) 
                     \end{bmatrix}
\end{equation}

As seen in Equation (\ref{eq:23}), the coordinate transformation yields the kinematic error model in the local coordinate frame. The rate of heading angle deviation can also be approximated by a small angle deviation around zero. These derivations are a simplified version of road-aligned (Frenet Frame) models given in \cite{carvalho2016predictive} by assuming small steering angle as well ($\delta - \delta_d = 0 $). In the road-aligned coordinate system, the heading error is expressed as $\dot{\Psi_e} = \dot{\psi}-\dot{\psi_d} =  \frac{V}{L}tan(\delta) -\kappa \dot{s}$ where the last term is the definition of desired heading rate and it acts as a disturbance input due to the curvature $\kappa$ in the equations. In designing feedback controllers, if we are not using prediction or disturbance compensation, this term can be neglected from the equations in feedback computations. The error in the heading angle can be simplified as given in Equation (\ref{eq:24}).  

\begin{equation}	
	\label{eq:24}
	\dot{\Psi_e} = \dot{\psi-\psi_d} = \frac{V}{L}tan(\delta-\delta_d)
\end{equation}

Combining Equations (\ref{eq:23}) and (\ref{eq:24}) in the matrix form results in the final kinematic error model.

\begin{equation}
\label{eq:25}
\begin{bmatrix}
	\dot{x}\\ \dot{y} \\ \dot{\psi_e}
\end{bmatrix}=\begin{bmatrix}
    	Vcos(\psi-\psi_d)\\
        Vsin(\psi-\psi_d) \\ 
        \frac{V}{L}tan(\delta-\delta_d)
	\end{bmatrix}
\end{equation}

if the kinematic model is linearized around the vicinity of a desired trajectory for which $$(s_*, u_*)=\{x=x_d, \:y=y_d, \:\psi_e=\psi-\psi_d=0 \:and\:\delta_e=\delta-\delta_d=0\}$$ and $$(s_e, u_e)=\{x-x_d, \:y-y_d, \:\psi_e, \:\delta_e \}$$

we obtain the linearized error equations in the local coordinate frame.

\begin{equation}
	\dot{s_e}=\dot{s-s_*}\:{\approx}\: \frac{\partial f(s, u)}{\partial s}\Bigr|_{\substack{s=s_*\\u=u_*}}(s-s_*)+\frac{\partial f(s, u)}{\partial u}\Bigr|_{\substack{s=s_*\\u=u_*}}(u-u_*)
\end{equation}

where the Jacobians are;

\begin{equation}
	\frac{\partial f(s, u)}{\partial s} = \begin{bmatrix}
	\frac{\partial (Vcos(\psi-\psi_d))}{\partial x}&\frac{\partial (Vcos(\psi-\psi_d))}{\partial y}& \frac{\partial (Vcos(\psi-\psi_d))}{\partial \psi}\\
    \frac{\partial (Vsin(\psi-\psi_d))}{\partial x}&\frac{\partial (Vsin(\psi-\psi_d))}{\partial y}& \frac{\partial (Vsin(\psi-\psi_d))}{\partial \psi}\\
    \frac{\partial (\frac{V}{L}tan(\delta-\delta_d))}{\partial x}&\frac{\partial (\frac{V}{L}tan(\delta-\delta_d))}{\partial y}& \frac{\partial (\frac{V}{L}tan(\delta-\delta_d))}{\partial \psi}
	\end{bmatrix}
\end{equation}

\begin{equation}
\frac{\partial f(s, u)}{\partial s} =\begin{bmatrix}
	0 &0 & -Vsin(\psi-\psi_d) \\ 0 &0 & Vcos(\psi-\psi_d) \\ 0 & 0 & 0 
	\end{bmatrix}_{\substack{\psi-\psi_d=0}}=\begin{bmatrix}
	0 & 0 & 0 \\ 0 & 0 & V \\ 0 & 0 & 0 
	\end{bmatrix}
\end{equation}

and

\begin{equation}
\frac{\partial f(s, u)}{\partial u} =\begin{bmatrix}
	\frac{\partial (Vcos(\psi-\psi_d))}{\partial \delta} \\ 
    \frac{\partial (Vsin(\psi-\psi_d))}{\partial \delta} \\
    \frac{\partial (\frac{V}{L}tan(\delta-\delta_d))}{\partial \delta}
	\end{bmatrix}_{\substack{\delta-\delta_d=0}}=\begin{bmatrix}
	0 \\ 0 \\ \frac{V}{L} 
	\end{bmatrix}
\end{equation}

The linear model for a constant vehicle speed $(V)$ is expressed as;

\begin{equation}
	\begin{bmatrix}
		\dot{x_e} \\ \dot{y_e} \\ \dot{\psi_e}
	\end{bmatrix} = \begin{bmatrix}
						0 & 0 & 0 \\
                        0 & 0 & V \\
                        0 & 0 & 0
					\end{bmatrix}
      \begin{bmatrix}
		  x_e \\ y_e \\ \psi_e
	  \end{bmatrix}+      
      \begin{bmatrix}
		  0 \\ 0 \\ \frac{V}{L}
	  \end{bmatrix}(\delta-\delta_d)      
\end{equation}

Since in this tutorial, the local longitudinal motion is not controlled, we can remove the longitudinal Equation from the model and proceed with the design of the Linear Quadratic Regulator (LQR) method for constant and time-varying vehicle speeds. The latter one is implemented by using a simple gain scheduling method at the grid of operating points.

In the case of constant vehicle speed, we can write the lateral model as;

\begin{equation}
	\begin{bmatrix}
		\dot{y_e} \\ \dot{\psi_e}
	\end{bmatrix} = \begin{bmatrix}
						0 & V \\
                        0 & 0
					\end{bmatrix}
      \begin{bmatrix}
		 y_e \\ \psi_e
	  \end{bmatrix}+      
      \begin{bmatrix}
		 0 \\ \frac{V}{L}
	  \end{bmatrix}(\delta-\delta_d)      
\end{equation}

Choosing the control with state feedback coefficients $(k_1, k_2)$ as $$\delta=k_1y_e+k_2\psi_e+\delta_d = \delta_{fb}+\delta_{ff}$$ gives us a combined state feedback; $\delta_{fb}=k_1y_e+k_2\psi_e$ and feed-forward; $\delta_{ff}=\delta_d$ controllers. The controlled system thus becomes;

\begin{equation}
	\label{eq:32}
	\begin{bmatrix}
		\dot{y_e} \\ \dot{\psi_e}
	\end{bmatrix} = \begin{bmatrix}
						0 & V \\
                        0 & 0
					\end{bmatrix}
      \begin{bmatrix}
		 y_e \\ \psi_e
	  \end{bmatrix}+      
      \begin{bmatrix}
		 0 \\ \frac{V}{L}
	  \end{bmatrix}(\delta_{fb})      
\end{equation}

\begin{figure}[H]
  \centering	
  \includegraphics[width=0.8\textwidth]{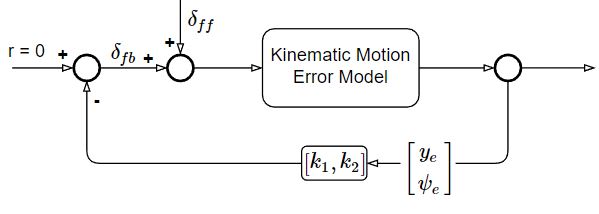}
  \captionsetup{justification=centering}
  \caption{Kinematic Controller Block Diagram}
  \label{fig:kin_block}
\end{figure}

The overall controller block diagram is shown in the Figure (\ref{fig:kin_block}).

It is trivial to design stabilizing LQR state space control coefficients $(k_1, \:k_2)$ for the given linear system and obtain time-varying controllers for different vehicle speed as $(k_1(V(t)), \:k_2(V(t)))$ in Matlab. In the following figure (Figure \ref{fig:gainscheduling}), we show the state feedback coefficients as the function of the vehicle speed ranging in $V(t)=[1, \:15]\:m/s$. The coefficients are obtained by giving equal weights to the states in the $Q$ and control effort $R$ matrices of the LQR controller. We discuss how to define the weights in $Q$ and $R$ matrices in the following sections. 

\begin{figure}[H]
  \centering	
  \includegraphics[width=1\textwidth]{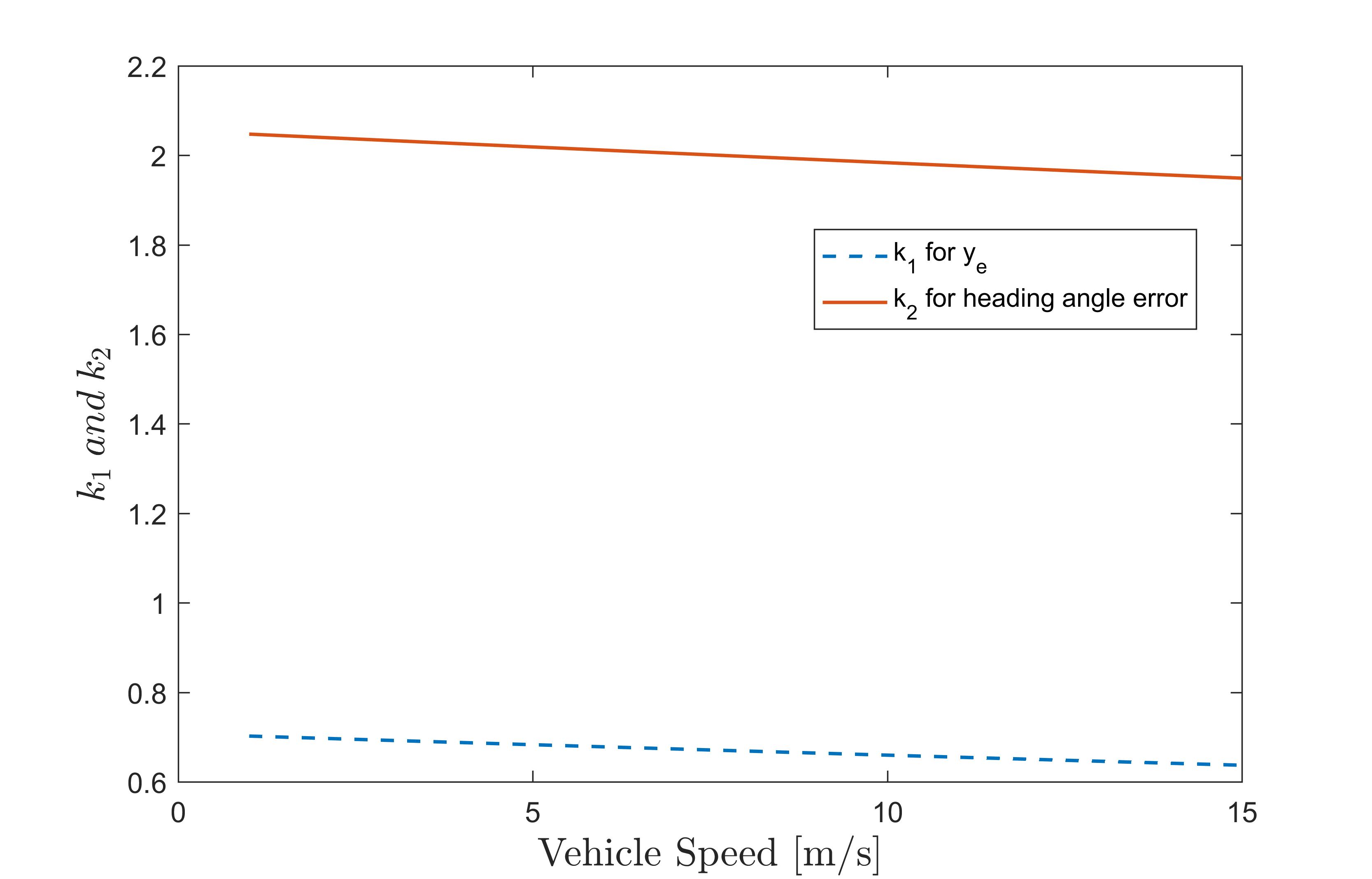}
  \captionsetup{justification=centering}
  \caption{Lateral and Heading Deviation error Feedback Gains vs Vehicle Speed}
  \label{fig:gainscheduling}
\end{figure}  

\paragraph{Feed-forward Control and Road Curvature}:\\[6pt]
We assumed that a trajectory generator provides the feed-forward part of the steering control. It can also be computed from the Ackerman geometry (Figure (\ref{fig:ackermann}) and Equation (\ref{eq:33})) if the path curvature ( Equation (\ref{eq:34})) is available. The Ackerman steering angle at tires is defined geometrically by the following equations;

\begin{align}
  \label{eq:33}
  \begin{split}
      	\delta_{ff} = atan(\frac{L}{R}) = atan({\kappa}L) 
  \end{split}\\
  \label{eq:34}
  \begin{split}
  		\kappa=\frac{1}{R}  
  \end{split} \\
  \label{eq:35}
  \begin{split}
  		\kappa=\frac{tan(\delta)}{L}  
  \end{split} 
\end{align}

The path curvature $(\kappa)$ path can be computed in various ways. The first method is to drive the car along a desired trajectory at low speeds and compute the curvature using Equation (\ref{eq:34} and \ref{eq:35}). These equations are only valid if the sideslip angle of the tires low. The second method to compute the road curvature from the measurements is to make use of the curvature differential equation \cite{jazar2017vehicle} given as; 

\begin{align}
	\
	\begin{split}
		\kappa = \frac{Y^{{\prime}{\prime}}}{\left( 1+ Y^{\prime}\right)^\frac{3}{2} }
	\end{split}
\end{align}

where $Y^{\prime}=\frac{dY}{dX}$ and $Y^{{\prime}{\prime}}=\frac{d^2Y}{dX^2}$  in the global coordinate system. The computation of the first and second derivatives numerically is always troublesome if there is no analytical and $C^2$ continuous function in the form of $Y=f(X)$. Alternatively, the Equation can be converted to a function of measurable variables. The first derivative of the coordinate (Y in the north direction) with respect to $X$  is related to the heading angle in the global coordinate system. 

\begin{equation}
\label{eq:37}
tan(\psi) = Y^{\prime}=\frac{dY}{dX}
\end{equation}

We can obtain $Y^{{\prime}{\prime}}$ by differentiating the Equation ({\ref{eq:37}}) with respect to $X$ and using the chain rule, we arrive at another equation of measurable variables.

\begin{align}
\label{eq:38}
    \begin{split}
        Y^{\prime}=\frac{dY}{dX} = tan(\psi)
    \end{split}\\
\label{eq:39}
    \begin{split}
        Y^{{\prime}{\prime}}=\frac{d}{dX}(\frac{dY}{dX}) =\frac{d}{dt}(\frac{dY}{dX})\frac{dt}{dX}
    \end{split} \\
\label{eq:40}
    \begin{split}
        Y^{{\prime}{\prime}}=\frac{d}{dt}(tan(\psi))\frac{dt}{dX}=\frac{\dot{\psi}}{{\mid}Vcos(\psi){\mid}cos(\psi)^2}
    \end{split}     
\end{align}

In Equation (\ref{eq:40}), $\frac{dX}{dt}$ is the vehicle speed on the $X~east$ direction in the global coordinate system and it is equivalent to $Vcos(\psi)$ and the derivative of $tan(\psi)$ is written as $\frac{\dot{\psi}}{cos(\psi)^2}$.

\begin{figure}[h]
  \centering	
  \includegraphics[width=0.6\textwidth]{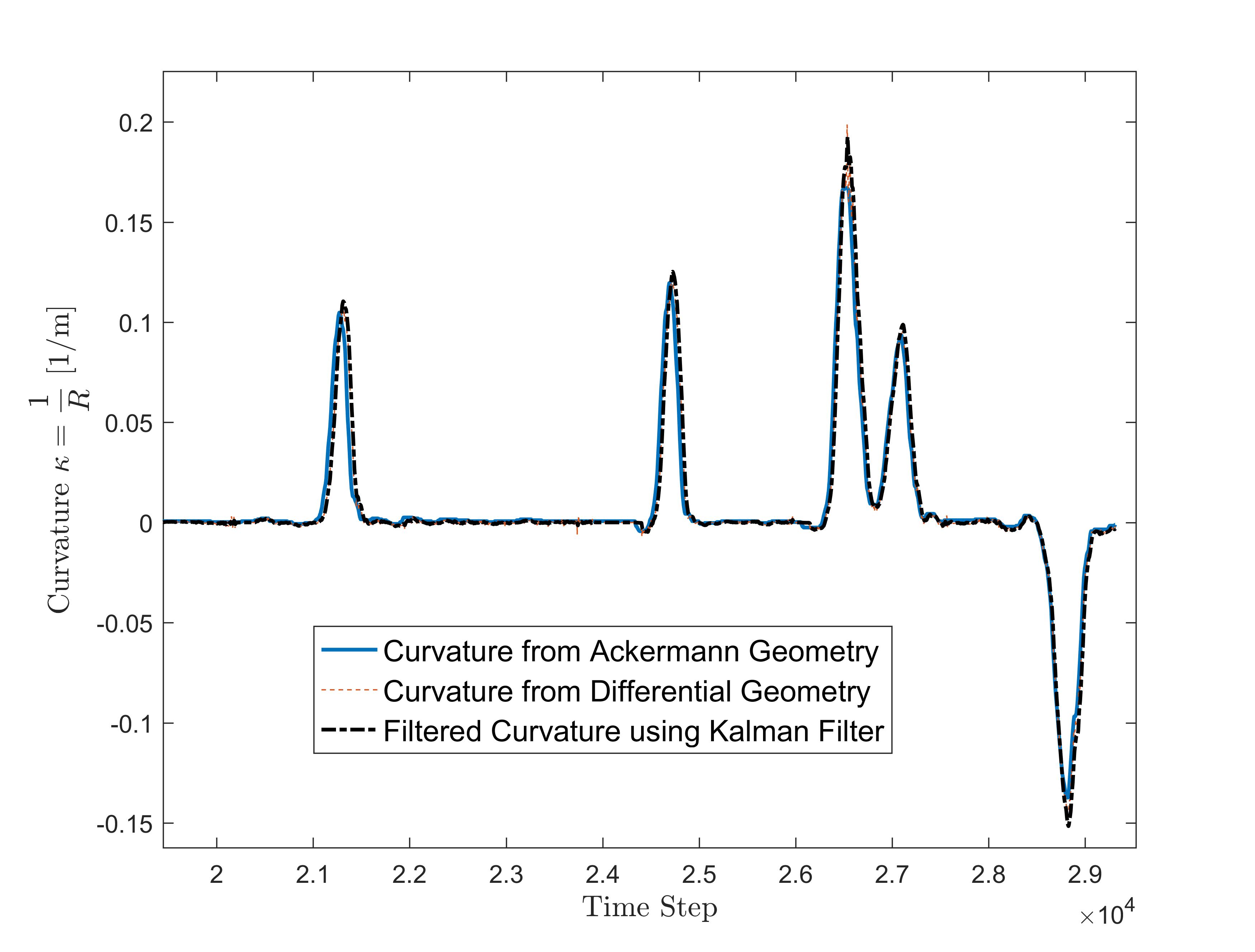}
  \captionsetup{justification=centering}
  \caption{Curvatures obtained by Ackermann Geometry, Differential Equations and Kalman Filtering}
  \label{fig:curvature}
\end{figure}

\begin{figure}[H]
  \centering	
  \includegraphics[width=0.6\textwidth]{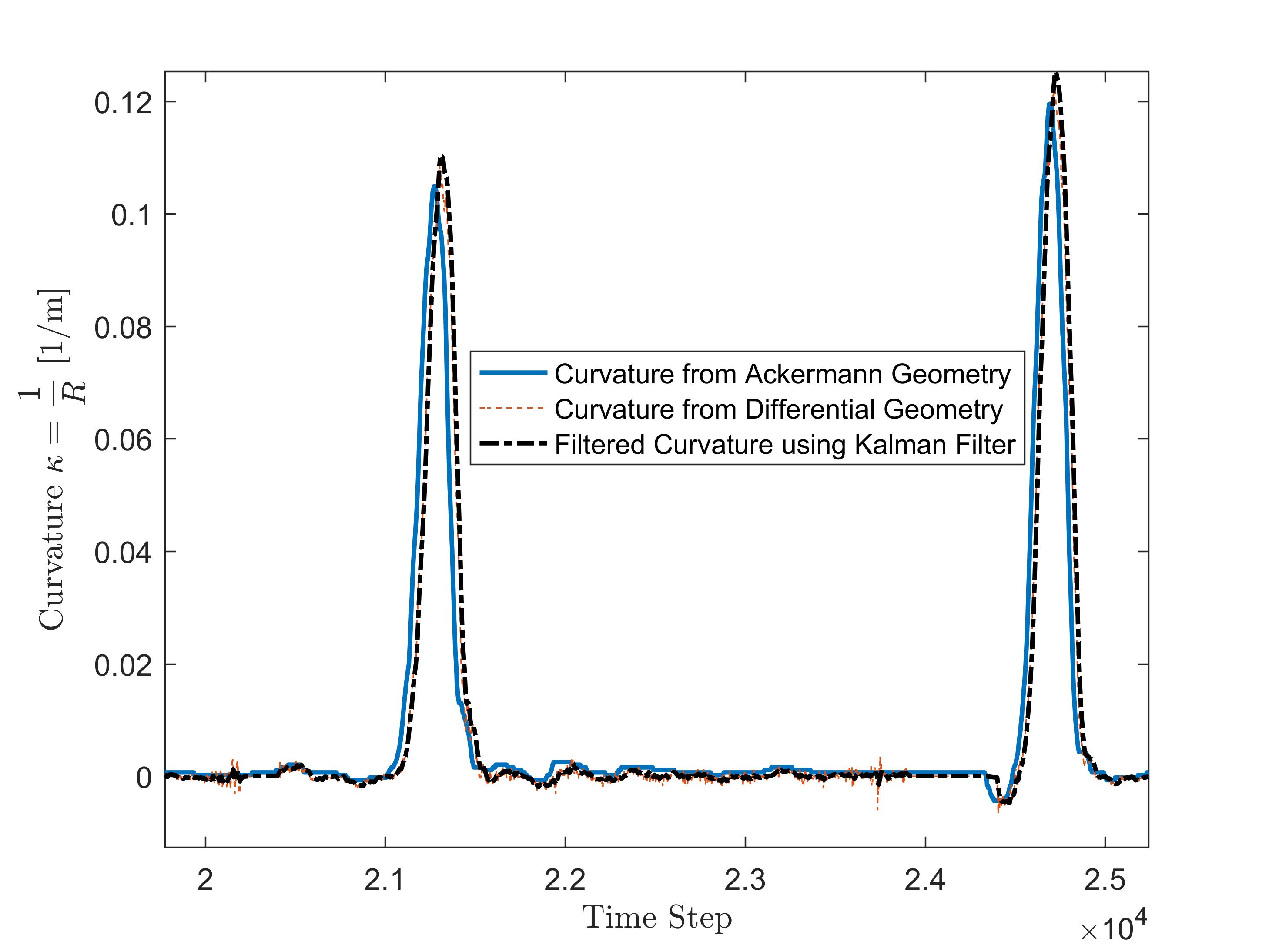}
  \captionsetup{justification=centering}
  \caption{Noise around zero line}
  \label{fig:curvature2}
\end{figure}

We derived the curvature function, which depends on the measurable quantities; heading angle, yaw rate and the vehicle velocity $\{\psi, \: \dot{\psi}\:and\:V\}$. As a result of these representations, a smooth curvature can be obtained from the collected data as shown in Figure \ref{fig:curvature}. A closer look around zero line is given in Figure (\ref{fig:curvature2}).  

As seen in the figures, both of the methods give almost identical results. The curvature computed using the Ackerman geometry and steering measurement has low resolution. The steering wheel angle is a low-resolution measurement obtained from the CAN bus. On the other hand, the curvature computed by the differential equations is noisy around the zero line due to the low vehicle speeds and high-frequency measurement (200 Hz) of the yaw rate. Furthermore, since the steering is the input and the vehicle response variables are the outputs, a time delay is observed between the input and output variables. This delay is a mechanical delay in the vehicle system and cannot be avoided. As we have two curvature measurements, we can also use a Kalman filter to obtain filtered curvature values. 

\begin{figure}[h]
  \centering	
  \includegraphics[width=1\textwidth]{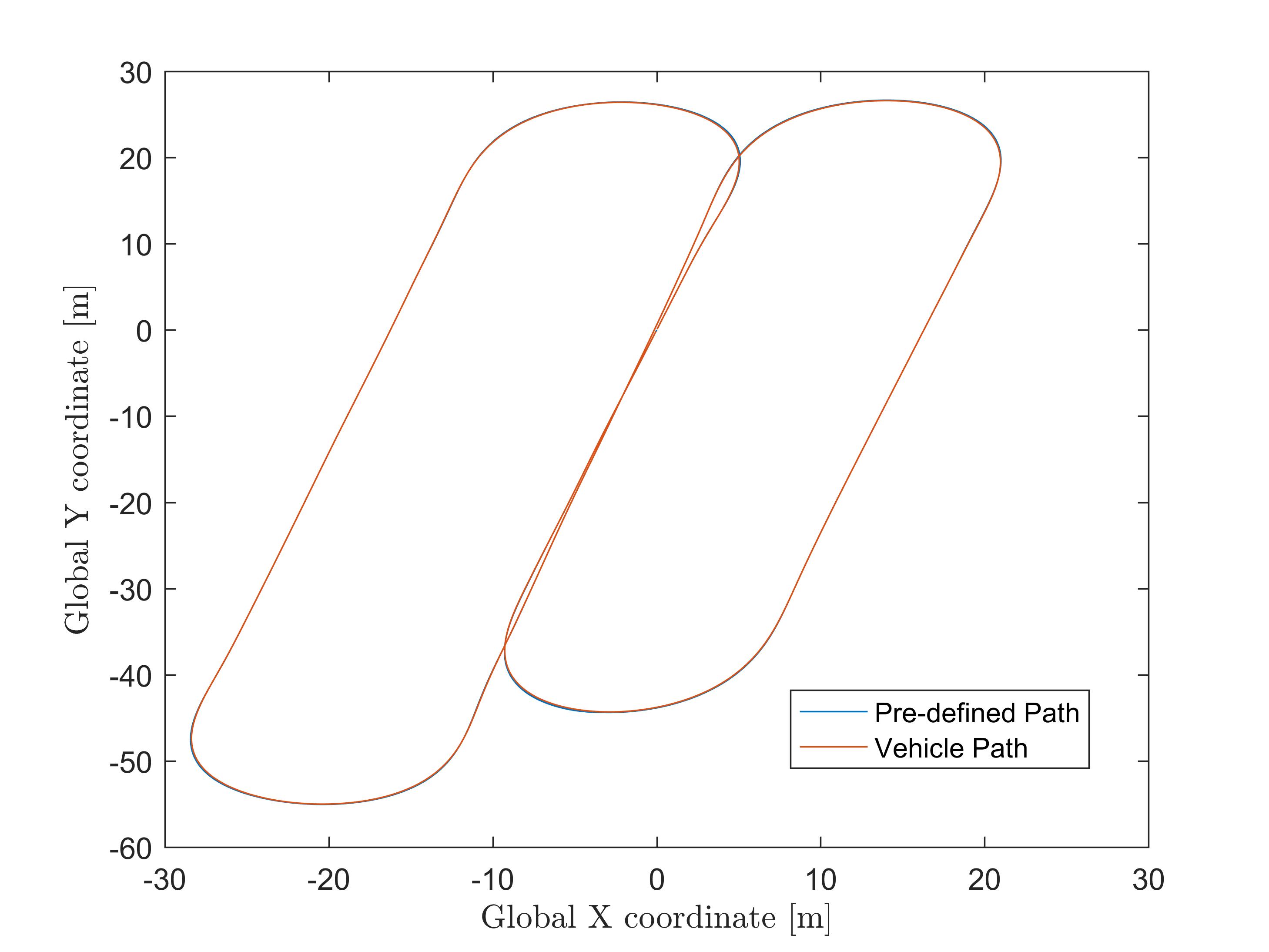}
  \captionsetup{justification=centering}
  \caption{Path Following Performance Simulation}
  \label{fig:rear_sim}
\end{figure}

\paragraph{Simulations and Experimental Results for Precise Parking}:

We simulated the controlled system to assess path following error under the given formulations for the rear and front axle equations. Figure (\ref{fig:rear_sim}) shows the simulated path and the path followed by the controlled vehicle with a constant speed of 10 [m/s]. The kinematic model with the controller can also be used to smooth the path obtained by the GPS measurements on the previously driven path. 

\begin{figure}[h]
  \centering	
  \includegraphics[width=1.0\textwidth]{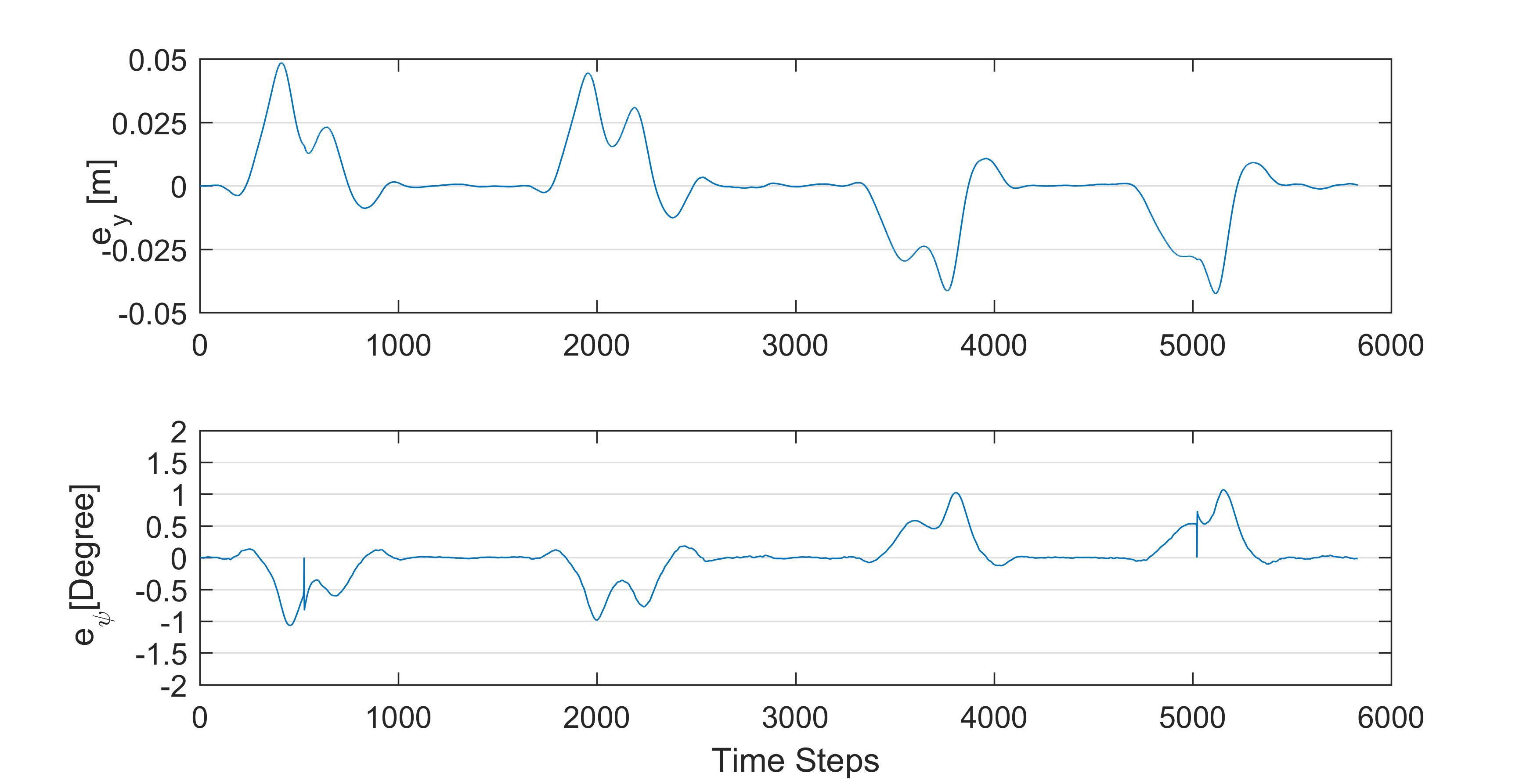}
  \captionsetup{justification=centering}
  \caption{Path Following Performance Simulation}
  \label{fig:tracking_error}
\end{figure}

The lateral error in the vehicle coordinate system is under 10 [cm] while turning with the constant speed of 10 [m/s]. The lateral tracking and heading errors are under five cm and one degree when the vehicle speed is three m/s (Figure \ref{fig:tracking_error}). 

\begin{figure}[H]
  \centering	
  \includegraphics[width=1.0\textwidth]{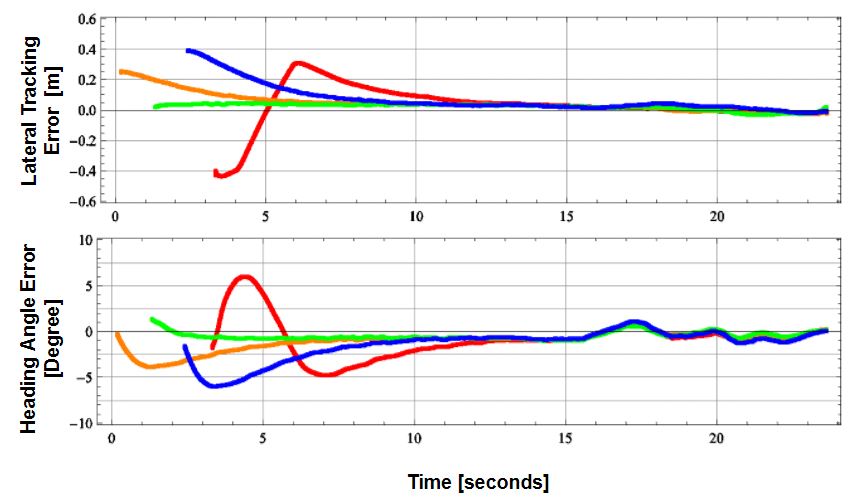}
  \captionsetup{justification=centering}
  \caption{Tight Parking Experiment Results starting from Various Positions}
  \label{fig:parking_exp}
\end{figure}

We verified the simulation results by employing the kinematic controller for tight parking application by starting from various positions away from the reference path. The reference path was obtained by driving the car from the parking slot to the final position. The experimental results are shown in Figure (\ref{fig:parking_exp}).

In the following section, we explain the essential ingredients of classical and modern control in an intuitive manner briefly before delving into the controller design for dynamical vehicle models. 

\subsection{Objectives and Methods in Control Theory }

\begin{figure}[H]
  \centering	
  \includegraphics[width=1.0\textwidth]{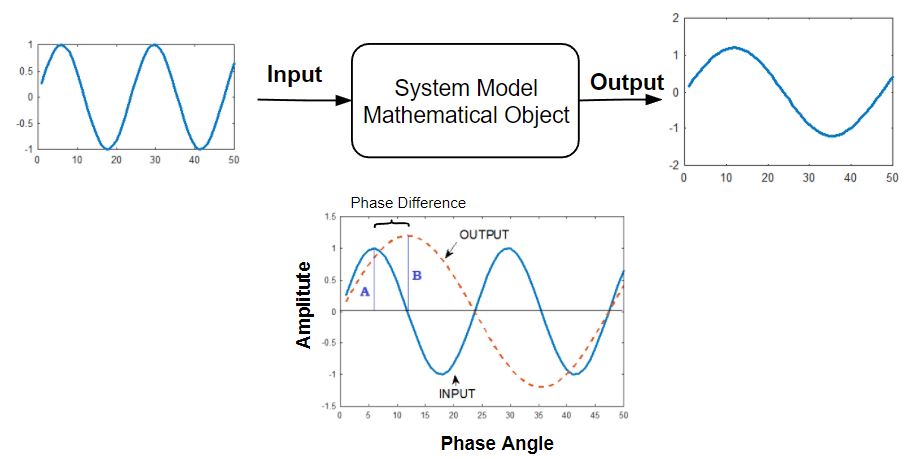}
  \captionsetup{justification=centering}
  \caption{Frequency Response}
  \label{fig:sininput}
\end{figure}

The system to be controlled is nothing but a mathematical object that takes the input signal and spits out the response signal, which we try to bring to the desired level in the control theory. The input signal is amplified or attenuated with some phase difference between their corresponding input and output values. We show this system behaviour in Figure (\ref{fig:sininput}). The system in the Figure takes a sinusoidal input with a magnitude value $A$, transforms it to another sinusoidal signal with a magnitude $B$ with a phase difference.  

In classical control, the time domain differential equations are converted to the frequency domain equations by the Laplace transformation to obtain a transfer function representation of the system. Exciting the system with inputs at different frequencies, the ratio of the input-output amplitudes as well as the corresponding phase angles are generated in the form of Bode plots. 

\begin{figure}[H]
  \centering	
  \includegraphics[width=1.0\textwidth]{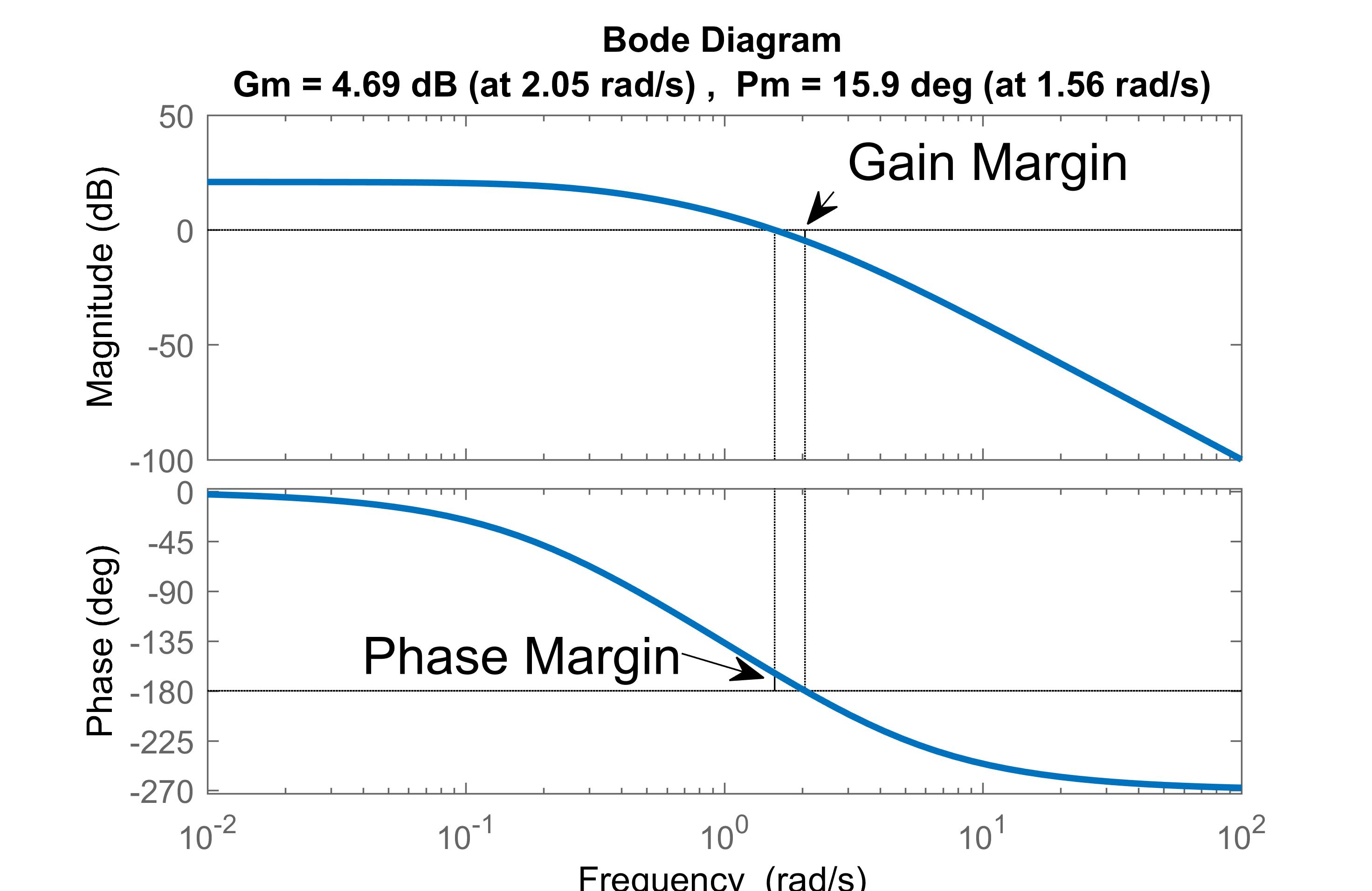}
  \captionsetup{justification=centering}
  \caption{Bode Diagram, Gain and Phase Margins}
  \label{fig:margins}
\end{figure}

The Bode plots are used to visualize the ratio of input-output magnitudes as well as the phase difference between them in a logarithmic scale. The unit of the magnitude ratio is decibel (dB). A typical Bode plot diagram for magnitudes and phase given in Figure (\ref{fig:margins}).

One of the pairs that is used for feedback stability assessment is gain and phase margins (also shown in Figure \ref{fig:margins} ). The term margin is used to indicate the allowable amount in changing some magnitude to the defined boundaries. The controllers change the system's phase and amplitude ratio. We need to define allowable magnitudes for multiplication and addition relative to the stability boundaries of the transfer function. The Nyquist stability criterion describes the stability of a transfer function in the Laplace domain. In simple terms, we can define a stable system among all other stability definitions, as the system produces Bounded Output for the Bounded Inputs (BIBO stability). There are other stability definitions. However, BIBO stability is sufficient to create some sense of stability notion as an introduction. 

\begin{figure}[H]
  \centering	
  \includegraphics[width=0.5\textwidth]{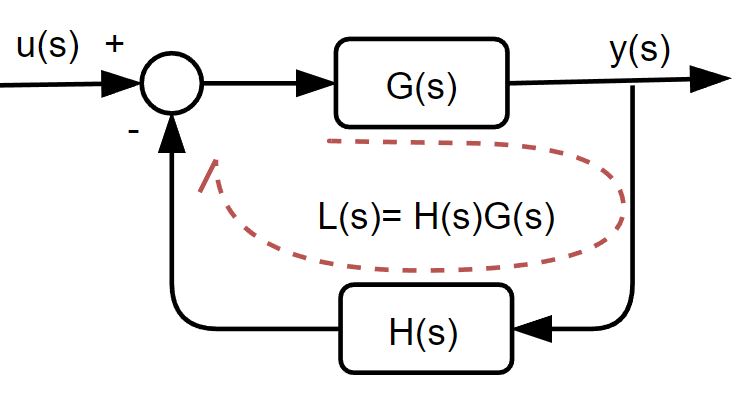}
  \captionsetup{justification=centering}
  \caption{Closed and Open Loop Transfer Functions}
  \label{fig:tf}
\end{figure}

The transfer function of a feedback control system given in Figure (\ref{fig:tf}) is written as; 

\begin{align}
  \begin{split}
		y(s) = \left[u(s)-y(s)H(s) \right] G(s)
  \end{split}\\
  \label{eq:42}
  \begin{split}
  		T(s) =\frac{y(s)}{u(s)} =\frac{G(s)}{1+H(s)G(s)}
  \end{split}	
\end{align}

The characteristic function of the transfer function $T(s)$ is the denominator term in Equation (\ref{eq:42}). The loop transfer function and the characteristic equations in these definition are  defined as $L(s)=H(s)G(s)$ and  $1+H(s)G(s) = 1+L(s)$ respectively. The transfer function becomes unstable when the denominator term goes zero ($T(s)=\frac{G(s)}{0} \rightarrow\infty$). Therefore, the stability boundaries, phase and gain margin can be defined accordingly. The formal definition of the instability condition is written as \cite{skogestad2007multivariable, doyle2013feedback, astrom2010feedback,maciejowski1989multivariable};

\begin{align}
	\begin{split}
		1+H(s)G(s) = 1 + L(s) = 0 \implies L(s)=-1
	\end{split}\\    
\end{align}

The loop transfer function is a complex number generator. The magnitude and phase angle of $|-1+j*0|$ are $|1|$ and $180^0$ 
In the classical control theory, the roots of the characteristic Equation of the transfer function define the system's stability behaviour. In the stable case, the roots of the characteristic function $1+L(s)$ must lie in the left half of the complex plane, and the magnitude of the loop transfer function must be less than one for all the possible frequencies if the magnitude of the loop transfer function is less than one ($|L(s)|<1$), the input signal is not amplified through the loop.  

The Gain Margin (GM) is defined at these conditions as the inverse of the gain that brings the amplitude of the system gain to one when the phase is $180^0$. 

$$(GM)|L(s)|\Bigr|_{\substack{\angle L(s)=180^0}}=1 \implies GM=\frac{1}{|L(s)|\Bigr|_{\substack{\angle L(s)=180^0}}}$$

The Phase Margin (PM) is conversely the amount of the phase angle required to bring the phase of the system's phase to  $180^0$ when the magnitude of the system gain is equal to one. 

$$PM = 180 + {\angle} L(s)\Bigr|_{\substack{|L(s)|=1}}$$

The gain margin tells the designer how much gain increases allowable until the instability boundaries. There are uncertainties in the physical models; therefore, we assume there might be unaccounted steady-state multiplication factors. The gain margin allows us to leave enough room for the unaccounted hidden gains in the control process. Generally, it is required to have $GM>2\:\approx\:6\:dB$ and $PM>30^0$ or more \cite{skogestad2007multivariable}. Although it is the sole indicator for robustness to the time delay, the phase margin provides room for time delays in the system. 

\subsection{Optimal Linear Quadratic Regulator (LQR) with Vehicle Dynamical Model for Path Tracking}

\paragraph{Optimal LQR State Feedback Controller}:

We assume the reader is familiar with LQR control and will not elaborate the derivation of the control structure using Algebraic Riccati and discrete counterpart for finite and infinite time cost functions. We refer the reader for a complete treatment of the optimal control and specifically LQR methods to the well-known reference books \cite{lqr1hespanha2009linear,lqr2anderson2007optimal,lqrbryson1975applied,lqrkirk2012optimal}.

The LQR methods are versatile in control applications from robotics to flight control due to the simplicity in implementation. We will design LQR state feedback controllers for lateral vehicle dynamics, starting with redefining the dynamical vehicle models in error coordinates as given in \cite{rajamani2011vehicle}. 

The controllers are designed either for continuous or discrete system models. The physical systems have a continuous response, however, the controllers work in a digital environment. Therefore, any controllers designed using continuous system equations must be discretized according to the control frequency to be used. Another approach is to design and use discrete controllers directly. In the rest of the paper, we will work with discrete systems. 

The discrete-time LQR state space control coefficients are derived using the Discrete Algebraic Riccati Equations. Assume that we have the following state-space equations in discrete time.

\begin{align}
\label{eq:45}
    x(k+1)  &= Ax(k)+Bu(k)   \\ 
    y(k) &= Cx(k) \nonumber
\end{align}

In Equation (\ref{eq:45}), $[x(k), \:u(k)]$ are the state and the control variables, $y(k)$ is the controlled outputs, $[A, \:, B, \:and\:C]$ are the state transition, control and measurement matrices respectively. We are after finding a state feedback control law of the form of $u(k)=Kx(k)$ with the state feedback coefficients $K$. The controller coefficients are derived from the solution of an optimal control problem to minimize the cost objective function $J_{LQR}$. For this purpose, in LQR optimal control framework, a quadratic cost is written, including weights for the controlled outputs and the control. Assuming that there is no constraint on states at the final time and the controller is to be derived for infinite time case for the steady-state solution. In this case, a quadratic cost function can be written as;

\begin{equation}
	J_{LQR} = \sum(y(k)^{T}\tilde{Q}y(k) + u(k)^{T}Ru(k))
\end{equation}

By substituting $y(k)=Cx(k)$ in the cost functional, we arrive at the cost function of the states;

\begin{equation}
	J_{LQR} = \sum(x(k)^{T}(C^{T}\tilde{Q})x(k) + u(k)^{T}Ru(k))
\end{equation}

where $Q=Q=C^{T}\tilde{Q}C{\succeq}0$ is the positive semidefinite and $R{\succ}0$ is positive definite matrices. For the control law of the form $u(k)=Kx(k)$, the state feedback control coefficients can be solved from the Discrete Algebraic Riccati matrix equation \cite{datta2004numerical,lancaster1995algebraic};

\begin{equation}
\label{eq:48}
	A^{T}XA-X+Q-A^{T}XB(R+B^{T}XB)^{-1}B^{T}XA=0
\end{equation}

where $X$ is the unique positive semidefinite steady-state solution to DARE (Equation \ref{eq:48}). Then, the state feedback coefficients are expressed as $K=(R+B^{T}XB)^{-1}A$. By substituting $u(k)=-Kx(k)$, we obtain the closed-form controlled state-space equation as $x(k+1)=(A-BK)x(k)$ where the closed loop state transition matrix is asymptotically stable and Hurwitz (all eigenvalues has negative real parts). 

\paragraph{Vehicle Lateral Dynamics in Error Coordinates}
LQR controller can be designed for in any coordinate system; however, for path tracking applications, it is more convenient to use the relative coordinate system \cite{rajamani2011vehicle}.  The vehicle dynamics model in the vehicle body coordinate system is given in matrix Equations (\ref{eq:17}). We define the desired lateral response by defining the desired magnitudes when vehicle is in steady-state motion conditions. The desired yaw rate and the acceleration in the steady-state turning are expressed as; 

\begin{align}
	\psi_d &=\frac{V_x}{R} \\ 
    \ddot{a}_{yd} &=\frac{V_x^2}{R} = V_x\dot{\psi_d}
\end{align}

where $R$ is the radius of curvature. The error in the lateral acceleration becomes;

$$\ddot{e}_y=a_y-a_{yd}=\ddot{y}+V_x\dot{\psi}-V_x\dot{\psi_d}=\ddot{y}+V_x(\dot{\psi}-\dot{\psi_d})$$

Similarly, the second error equation is written as;

$$\ddot{e}_{\psi}=\ddot{\psi}-\ddot{\psi_d}$$

If we substitute the equations given for $\ddot{y}$ and $\dot{\psi}$ in Equations (\ref{eq:14} and \ref{eq:16}) in the error equations, we obtain the error state space model given as;

\begin{multline}
\label{eq:51}
\frac{d}{dt}
	\begin{bmatrix}
	e_y \\ \dot{e_y} \\ e_\psi \\ \dot{e_\psi}
	\end{bmatrix} = 
    \begin{bmatrix}
    0 & 1 & 0 & 0 \\
    0 & -\frac{2 C_{\alpha f}+2 C_{\alpha r}}{m V_{x}} & \frac{2 C_{\alpha f}+2 C_{\alpha r}}{m V_{x}} &\frac{-2 C_{\alpha f} \ell_{f}+2 C_{\alpha r} \ell_{r}}{m V_{x}}\\
    0 & 0 & 1 & 0 \\
    0 &-\frac{2 C_{\alpha f} \ell_{f}-2 C_{\alpha r} \ell_{r}}{I_{z} V_{x}}&  \frac{2 C_{\alpha f} \ell_{f}-2 C_{\alpha r} \ell_{r}}{I_{z}}  &-\frac{2 C_{\alpha f} \ell_{f}^{2}+2 C_{\alpha r} \ell_{r}^{2}}{I_{z} V_{x}}
    \end{bmatrix}
    \begin{bmatrix}
		e_y \\ \dot{e_y} \\ e_\psi \\ \dot{e_\psi}
	\end{bmatrix}+
    \begin{bmatrix}
	0 \\ \frac{2C_{{\alpha}f}}{m}\\ 0 \\ \frac{2{L_f}C_{{\alpha}f}}{I_z}
	\end{bmatrix} \delta \\ +     \begin{bmatrix}
		0 \\-\frac{2 C_{\alpha f} \ell_{f}-2 C_{\alpha r} \ell_{r}}{m V_{x}}-V_{x} \\0 \\-\frac{2 C_{\alpha f} \ell_{f}^{2}+2 C_{\alpha r} \ell_{r}^{2}}{I_{z} V_{x}}
	\end{bmatrix} \dot{\psi_{des}} 
\end{multline}

Based on the state update equation (\ref{eq:51}) \cite{rajamani2011vehicle} and measurement matrix, we can design LQR controllers both for continuous and discrete cases. Usually, path deviation, lateral velocity, heading angle, and rate can be measured or derived in the control applications. 

\subsection{Path Generation}

The reference path must be available in real-time to the lateral motion controller. It can be computed in real-time by using numerical optimal control and trajectory optimization methods or prepared in advance if there is no need to change the path during the motion.  For parking applications, if the path is to be prepared off-line, the curvature of the path is necessary for the feed-forward part of the kinematic controller presented in this study. We provided different curvature computations in the previous sections. 
 
 \section{Conclusion}
In this tutorial, we briefly introduced practical control methods for autonomous vehicle steering applications. The kinematic and dynamic controller are described with experimental results. In reality, the control requirements are tighter than presented in this introductory tutorial. Various modules for controllers must be implemented with robust control theory to meet the control demands for autonomous driving. We proposed self-scheduling preview robust controllers for autonomous driving in \cite{boyalinagoya}. In addition to the robust controllers, robust observers and predictors must be used for sensor measurements. In the autonomous vehicle control applications, the path generation and planning are implemented separately, and the generated paths are sent to the low-level controllers. 

\section{Example Codes}
We provided accompanying codes with this paper. 

\nocite{*}
\printbibliography 
\end{document}